\documentclass[letterpaper, 10 pt, conference]{ieeeconf}  

\IEEEoverridecommandlockouts                              

\usepackage{amsmath} 
\usepackage{amssymb}  
\let\labelindent\relax 
\usepackage{enumitem}
\usepackage{graphicx}
\usepackage{xcolor}
\usepackage{balance}
\usepackage{booktabs} 
\usepackage{caption}  
\usepackage{subcaption}
\usepackage{url}
\usepackage[normalem]{ulem}

\title{\LARGE \bf
WM-DAgger: Enabling Efficient Data Aggregation for Imitation Learning with World Models
}

\author{Anlan Yu$^{1\dagger}$, Zaishu Chen$^{2\dagger}$, Peili Song$^{3}$, Zhiqing Hong$^{4*}$, Haotian Wang$^{2}$,\\Desheng Zhang$^{5}$, Tian He$^{2}$, Yi Ding$^{6}$, and Daqing Zhang$^{1,7*}$
\thanks{$^{\dagger}$These authors contributed equally. $^{*}$Corresponding authors.}%
\thanks{$^{1}$Anlan Yu is with Peking University, Beijing, China. (E-mail: yal6040@pku.edu.cn)}%
\thanks{$^{2}$Zaishu Chen, Haotian Wang, and Tian He are with JD Logistics, China. (Email: \{chenzaishu.1, wanghaotian18, tim.he\}@jd.com)}%
\thanks{$^{3}$Peili Song is with Nankai University, Tianjin, China. (Email: peilisong@mail.nankai.edu.cn)}%
\thanks{$^{4}$Zhiqing Hong is with The Hong Kong University of Science and Technology (Guangzhou), China. (E-mail: zhiqinghong@hkust-gz.edu.cn)}%
\thanks{$^{5}$Desheng Zhang is with Rutgers University, Piscataway, USA. (E-mail: desheng@cs.rutgers.edu)}%
\thanks{$^{6}$Yi Ding is with the University of Texas at Dallas, Richardson, TX, USA. (E-mail: yi.ding@utdallas.edu)}%
\thanks{$^{1,7}$Daqing Zhang is with Peking University, Beijing, China, and Institut Polytechnique de Paris, France. (E-mail: dqzhang@sei.pku.edu.cn)}%
}

\begin{document}

\maketitle
\thispagestyle{empty}
\pagestyle{empty}

\begin{abstract}


Imitation learning is a powerful paradigm for training robotic policies, yet its performance is limited by compounding errors: minor policy inaccuracies could drive robots into unseen out-of-distribution (OOD) states in the training set, where the policy could generate even bigger errors, leading to eventual failures.
While the Data Aggregation (DAgger) framework tries to address this issue, its reliance on continuous human involvement severely limits scalability. 
In this paper, we propose WM-DAgger, an efficient data aggregation framework that leverages World Models to synthesize OOD recovery data without requiring human involvement.
Specifically, we focus on manipulation tasks with an eye-in-hand robotic arm and only few-shot demonstrations.
To avoid synthesizing misleading data and overcome the hallucination issues inherent to World Models, our framework introduces two key mechanisms: (1) a Corrective Action Synthesis Module that generates task-oriented recovery actions to prevent misleading supervision, and (2) a Consistency-Guided Filtering Module that discards physically implausible trajectories by anchoring terminal synthesized frames to corresponding real frames in expert demonstrations. We extensively validate WM-DAgger on multiple real-world robotic tasks. Results   that our method significantly improves success rates, achieving a 93.3\% success rate in soft bag pushing with only five demonstrations. The source code is publicly available at \url{https://github.com/czs12354-xxdbd/WM-Dagger}.

\end{abstract}

\section{INTRODUCTION}




Imitation learning is a powerful paradigm for transferring human expertise to robotic systems~\cite{schaal1999imitation, zare2024survey, bai2025towards}. However, its effectiveness is hindered by \textit{compounding errors}\cite{ross2011reduction}: Even minor policy inaccuracies can drive the robot into states not covered by human demonstrations (i.e., out-of-distribution~(OOD) states), where the policy generates further errors, eventually leading to task failure.


To address this problem, the Data Aggregation (DAgger) paradigm~\cite{ross2011reduction, kelly2019hg} relies on continuous human intervention to guide the robot back from OOD states while simultaneously collecting recovery data for further imitation learning. However, its dependence on manual operation limits its scalability in practice.
Recently, diffusion-based models have been used to synthesize OOD recovery data~\cite{zhang2024diffusion}. However, their restriction to single-frame generation prevents them from modeling continuous dynamics of recovery processes. Moreover, without inherent understanding of environment dynamics, these models struggle with complex physical interactions and deformable objects. 

Recent advancements in World Models (WMs)~\cite{ding2025understanding, li2025simworld, li2025comprehensive} offer a unique opportunity to address this compounding error problem. World Models, such as Cosmos-Predict~\cite{agarwal2025cosmos}, take historical frames and intended actions as inputs to synthesize the resulting frames of future states. By synthesizing OOD recovery data with these models, the human effort required for data collection in the DAgger paradigm can be substantially reduced. Moreover, unlike existing single-frame diffusion models for data synthesis, WMs can generate multiple continuous frames that capture environment dynamics across complex real-world scenarios~\cite{ding2025understanding}, making them suitable for tasks involving complex physical interactions and deformable objects.


\begin{figure}[t]   
    \centering
    \includegraphics[width=0.45\textwidth]{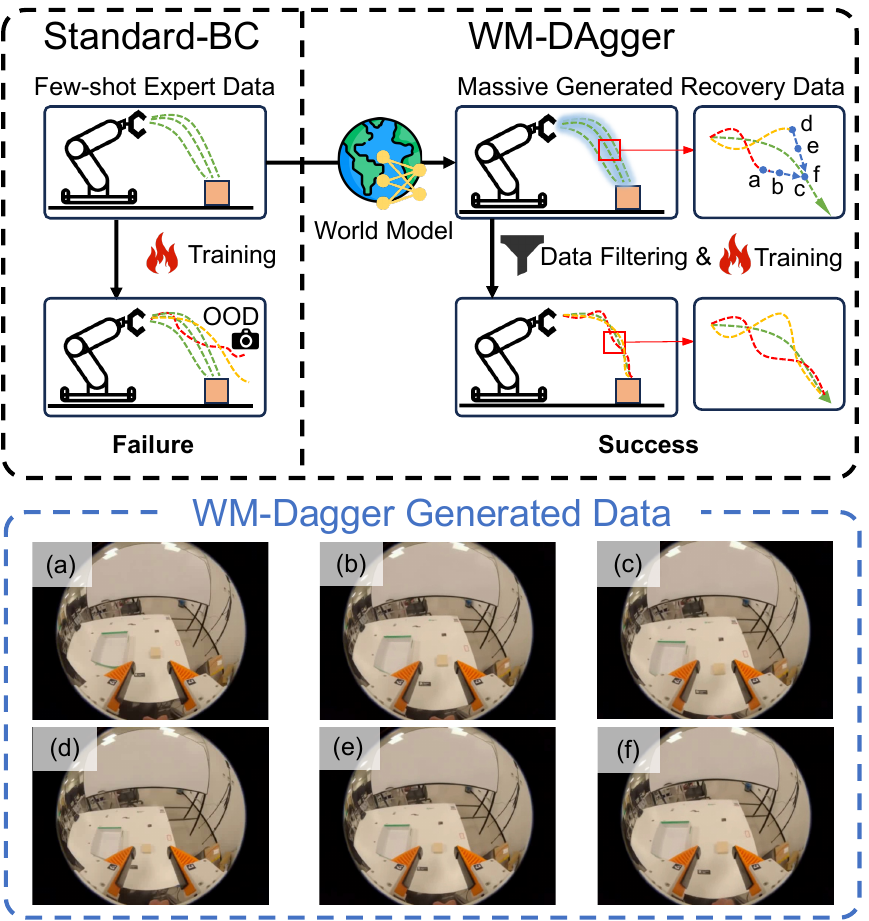}
    \caption{\textbf{WM-DAgger} mitigates the compounding errors of standard Behavioral Cloning (BC) by generating massive recovery supervision with a world model. (e.g., visual transitions $a \to b \to c$ and $d \to e \to f$). }
    \label{fig:teaser} 
    \vspace{-1em}
\end{figure}

Although promising, leveraging WMs to synthesize high-quality OOD recovery data is non-trivial.
\textbf{The first challenge} is the absence of expert involvement to provide optimal recovery supervision. While standard human-in-the-loop DAgger~\cite{kelly2019hg} relies on experts to supply OOD recovery actions, this manual labeling is unavailable when synthesizing trajectories within a world model. If these OOD states are paired with suboptimal actions, it will severely mislead policy training~\cite{brown2019extrapolating}.
\textbf{The second challenge} lies in mitigating the impact of hallucinations within the synthesized recovery data. WMs inevitably introduce hallucinations, such as object morphing or implausible state transitions. Training a policy on such data yields marginal performance improvements, as the policy learns from erroneous physical dynamics that mismatch the real-world environment.



In this work, we propose \textbf{WM-DAgger}, a framework leveraging \uline{W}orld \uline{M}odels to enable efficient \uline{D}ata \uline{Agg}regation for imitation learning. Specifically, we focus on robotic manipulation tasks utilizing the widely adopted eye-in-hand configuration~\cite{zhang2024diffusion, chi2024universal}. As illustrated in Figure~\ref{fig:teaser}, our framework trains the policy using few-shot expert demonstrations augmented by large-scale OOD recovery data, which is synthesized by the proposed Eye-in-Hand-Conditioned World Model~(EAC-WM) based on Cosmos-Predict-2.5~\cite{agarwal2025cosmos}. The model maps robotic actions into pixel-wise motion fields representing eye-in-hand camera displacement, providing explicit geometric conditioning to synthesize visual observations corresponding to the recovery actions.
\textbf{To address the first challenge,} we introduce a Corrective Action Synthesis Module. Instead of generating random recovery actions that could provide misleading supervision, this module synthesizes task-oriented recovery actions. Synthesized actions are constrained to follow the original direction of the expert trajectory. This prevents the policy from learning suboptimal actions that could even degrade performance.
\textbf{To address the second challenge,} we design a Consistency-Guided Filtering Module. It compares each synthesized OOD recovery data's terminal frame with the frame from the original expert data. As WMs inherently accumulate errors over time~\cite{li2025comprehensive}, the terminal frame contains the highest level of potential hallucination. If this frame remains visually and physically consistent with the real expert frame, the whole synthesized data is validated for policy training.


In summary, our contributions are as follows.
\begin{itemize}[leftmargin=*]
    \item Conceptually, we first utilize World Models to synthesize OOD recovery data, enabling robust imitation learning even from few-shot demonstrations.

    \item Technically, we propose WM-DAgger, a world-model-based OOD recovery data synthesis framework that generates task-oriented data with suppressed hallucinations.
    
    \item Experimentally, real-world evaluations across multiple tasks confirm the utility of WM-DAgger. Notably, it achieves a 93.3\% success rate in soft bag pushing using only five demonstrations.
\end{itemize}
\section{RELATED WORK}
\subsection{Dataset Aggregation for Imitation Learning}
Imitation learning is a vital paradigm for training robotic policies, yet it is often limited by distribution shifts and compounding execution errors in real-world settings~\cite{schaal1999imitation, zare2024survey, bai2025towards}. The Dataset Aggregation (DAgger) framework~\cite{ross2011reduction} addresses this by iteratively incorporating expert feedback in recovering from OOD states to train the policy. While variants like HG-DAgger~\cite{kelly2019hg} and CR-DAgger~\cite{xu2025compliant} optimize this process through controlled gating and compliant intervention, they remain constrained by a fundamental reliance on expensive human-in-the-loop supervision. 

To enhance scalability, recent generative approaches such as DMD~\cite{zhang2024diffusion} synthesize OOD data for imitation learning; however, these methods often prioritize visual synthesis and lack physical awareness in the synthesized samples. In this work, we propose WM-DAgger, which utilizes a world model to autonomously synthesize physically consistent corrective trajectories, enabling robust imitation learning without additional human cost for data aggregation.


\subsection{World Models in Robot Learning}

World Models serve as internal simulators of the environment, enabling agents to predict the causal consequences of their actions~\cite{ding2025understanding, li2025simworld, yu2025manigaussian++}. 
One prominent approach, exemplified by the Dreamer series~\cite{hafner2023mastering}, encodes high-dimensional observations into a compact latent space using a Recurrent State-Space Model. 
Recent advancements have introduced powerful generative world models trained on internet-scale data. Systems like NVIDIA Cosmos~\cite{agarwal2025cosmos} utilize video diffusion to simulate realistic physical interactions based on language instructions or motor controls, achieving strong generalization capabilities.
In practical robotics, frameworks such as DayDreamer~\cite{wu2023daydreamer} demonstrate that world models facilitate rapid learning on physical robots without human demonstrations. Similarly, World4RL~\cite{jiang2025world4rl} improves manipulation success rates by training policies on imagined rollouts, thereby avoiding the safety risks of real-world trials.
In this work, we leverage the strong generalization capability of pre-trained world models to address the distribution shift problem in imitation learning.
\begin{figure*}[t]   
    \centering
    \includegraphics[width=1\textwidth]{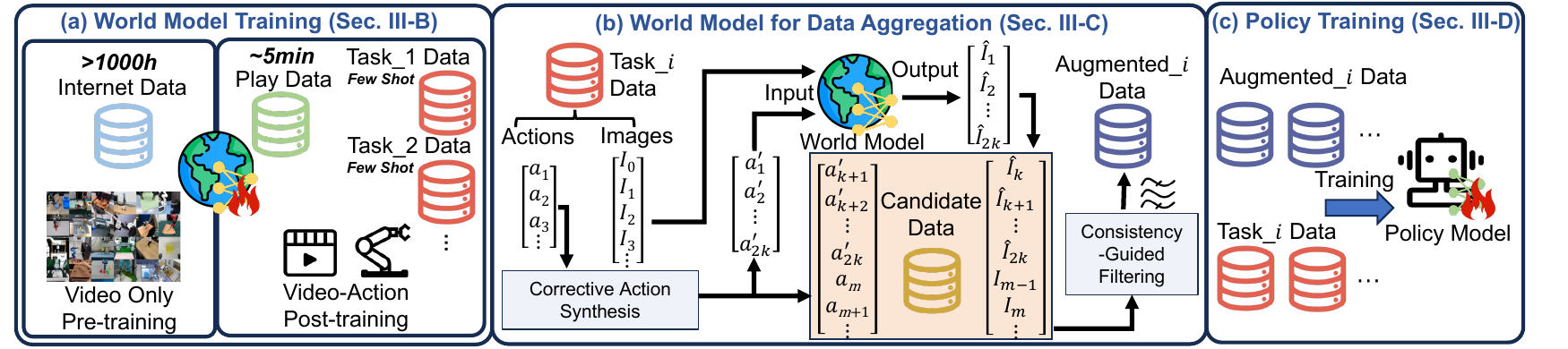}
    \caption{Overall pipeline of our WM-DAgger framework.}
    \label{fig:pipeline}
    \vspace{-1em}
\end{figure*}

\section{METHODOLOGY}

We introduce WM-DAgger, a world-model-driven framework that synthesizes OOD states and corresponding recovery trajectories to train robust policies. The overall pipeline, illustrated in Figure~\ref{fig:pipeline}, begins by training an eye-in-hand, action-conditioned world model. 
It learns physically consistent environmental dynamics using a combination of explorative Play Data and expert-demonstrated Task Data. Specifically, Play Data is collected via goal-agnostic exploration of the workspace, while Task Data comprises specific expert trajectories to accomplish targeted manipulation tasks.
Utilizing this trained WM, we generate large-scale OOD recovery data from the expert Task Data. Specifically, the Corrective Action Synthesis Module produces recovery actions by intentionally deviating from the expert trajectory into surrounding OOD states and subsequently steering back. The WM then synthesizes the corresponding visual frames along these recovery paths. To ensure data fidelity, the Consistency-Guided Filtering Module discards trajectories that exhibit physically implausible hallucinations in the synthesized frames. Finally, we aggregate the original expert Task Data with the validated synthetic recovery data to train a policy capable of rectifying compounding execution errors.

\subsection{Eye-in-Hand Action-Conditioned World Model Design}
\begin{figure}[ht]   
    \centering
    \includegraphics[width=0.46\textwidth]{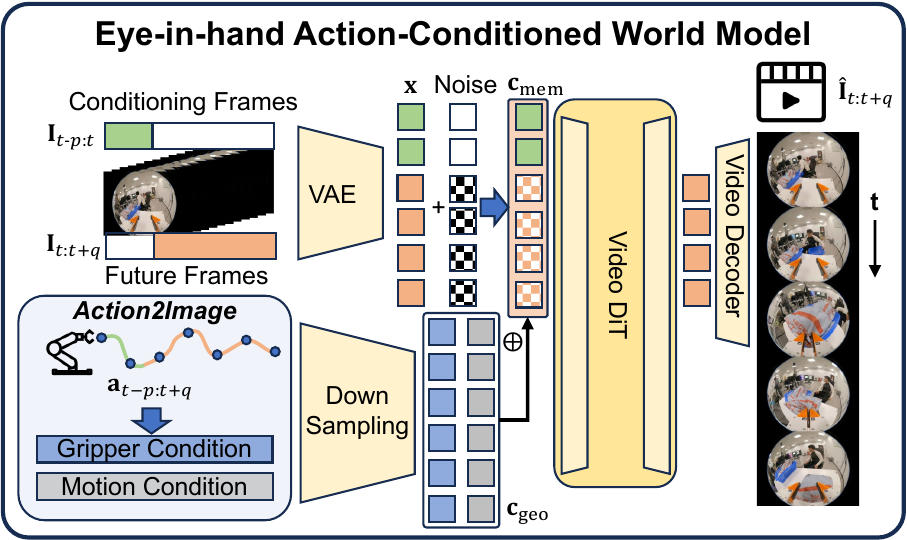}
    \caption{EAC-WM Architecture.}
    \label{fig:wm_arch}
    \vspace{-1em}
\end{figure}




\subsubsection{Overall Architecture}
\label{subsec:model_architecture}

We propose the \textbf{Eye-in-Hand Action-Conditioned World Model (EAC-WM)}, an architecture designed to capture and synthesize eye-in-hand visual dynamics. 
Built upon the GE-Sim~\cite{liao2025genie} framework with a Cosmos-Predict2.5 (2B)~\cite{agarwal2025cosmos} backbone, EAC-WM introduces an \textit{Action2Image} conditioning module. By translating actions into the relative spatial movement of each pixel in the eye-in-hand camera, this module conditions the WM with fine-grained geometric information, thereby improving the fidelity of predicted eye-in-hand visual dynamics.


We define the robotic control input at time $t$ as a vector $\mathbf{a}_t = [\mathbf{t}_t, \mathbf{q}_t, g_t]^\top$, comprising the Cartesian translation $\mathbf{t}_t \in \mathbb{R}^3$, unit quaternion orientation $\mathbf{q}_t \in \mathbb{R}^4$, and continuous gripper state $g_t \in \mathbb{R}$. Formally, let $f_\theta$ denote the WM parameterized by $\theta$. Given a context window of $p$ historical observations and a sequence of $q$ future actions, EAC-WM predicts the subsequent $q$ visual states:
\begin{equation}
    \hat{\mathbf{I}}_{t:t+q} = f_\theta \left( \mathbf{I}_{t-p:t}, \mathbf{a}_{t-p:t+q} \right),
\end{equation}
where $\mathbf{I}$ and $\hat{\mathbf{I}}$ represent the ground-truth and synthesized visual frames, respectively.


\subsubsection{Action2Image Conditioning}

Conditioning the world model $f_\theta$ on low-dimensional  actions $\mathbf{a}_{t-p:t+q}$ is challenging, as sparse action vectors are easily overshadowed by high-dimensional visual conditions $\mathbf{I}_{t-p:t}$. Inspired by camera parameter encoding methods in computer vision~\cite{yifan2022input}, we introduce the \textit{Action2Image} module, which projects sparse action vectors into a dense, pixel-aligned geometric condition to bridge this dimensional gap. 



In this module, we first derive the camera's rotation $\mathbf{R}_t \in SO(3)$ and translation $\mathbf{o}_t \in \mathbb{R}^3$ in the world coordinate system by transforming the action vector $[\mathbf{t}_t, \mathbf{q}_t]$ via a fixed hand-eye calibration matrix. Based on 3D projective geometry, each image pixel $(u, v)$ is associated with a 3D ray originating at the camera's optical center. The unit viewing direction $\mathbf{d}_t^{(u,v)}$ of the pixel $(u, v)$ can be computed using the camera intrinsic matrix $\mathbf{K} \in \mathbb{R}^{3 \times 3}$ as:
\begin{equation}
\mathbf{d}_t^{(u,v)} = \frac{\mathbf{R}_t \mathbf{K}^{-1}[u, v, 1]^\top}{\left\| \mathbf{R}_t \mathbf{K}^{-1}[u, v, 1]^\top \right\|_2}.
\end{equation}

To effectively condition the model to generate the image at step $t+i$ based on the image at step $t$ and the relative action between them, we transform the relative action into high-dimensional relative ray transformations for each pixel in the image. The relative translation $\Delta\mathbf{o}_{t+i} = \mathbf{o}_{t+i} - \mathbf{o}_t$ is spatially broadcasted across the image dimensions to form a uniform origin displacement grid $\Delta\mathbf{O}_{t+i} \in \mathbb{R}^{H \times W \times 3}$. Simultaneously, we compute the per-pixel directional shift $\Delta\mathbf{d}_{t+i}^{(u,v)} = \frac{\mathbf{d}_{t+i}^{(u,v)} - \mathbf{d}_t^{(u,v)}}{\left\| \mathbf{d}_{t+i}^{(u,v)} - \mathbf{d}_t^{(u,v)} \right\|_2}$, which is aggregated into a dense directional tensor $\Delta\mathbf{D}_{t+i} \in \mathbb{R}^{H \times W \times 3}$. By concatenating $\Delta\mathbf{O}_{t+i}$ and $\Delta\mathbf{D}_{t+i}$, we acquire motion condition.
To incorporate non-spatial action condition, the scalar gripper state $g_{t+i}$ is broadcasted into a feature map $\mathbf{C}_{\text{grip}} \in \mathbb{R}^{H \times W \times 1}$ as the gripper condition. These components are concatenated to construct the final dense geometric condition:
\begin{equation}
\mathbf{C}_{\text{geo}, t+i} = \big[ \Delta\mathbf{O}_{t+i}, \, \Delta\mathbf{D}_{t+i}, \, \mathbf{C}_{\text{grip}} \big] \in \mathbb{R}^{H \times W \times 7}.
\end{equation}

\subsubsection{Video-Action Tokenization}
To process high-dimensional visual and geometric data with world foundation model, we project inputs into a compact latent space using a pretrained Variational Autoencoder (VAE). Given a sequence of historical context frames $\mathbf{I}_{t-p:t}$ and target future frames $\mathbf{I}_{t:t+q}$, the VAE yields visual image tokens $\mathbf{x}_{t-p:t+q}$. Simultaneously, the dense geometric condition $\mathbf{C}_{\text{geo}}$ is projected through down-sampling to obtain action tokens $\mathbf{c}_{\text{geo}}$. This unified tokenization ensures that both the visual observations and the robotic control actions are represented in a shared latent space, facilitating effective cross-modal interaction within the world foundation model.


\subsubsection{World Model Training and Data Strategy}
EAC-WM is initialized with the Cosmos-Predict2.5 (2B) foundation model, which establishes general physical priors through pre-training on internet-scale video-only data $\mathcal{D}_I$. We subsequently adapt the model to specific manipulation dynamics using synchronized visual-action sequences from Play Data $\mathcal{D}_P$ and Task Data $\mathcal{D}$. Play data $\mathcal{D}_P$ consists of unscripted human exploratory movements used to internalize scene-specific geometry, while Task data $\mathcal{D}$ comprises expert demonstrations that refine the model's understanding of action-conditioned physical dynamics.

The training of EAC-WM is formulated within the Rectified Flow framework~\cite{liu2022flow}, which learns to map noise to data along a deterministic linear path. 
During our video-action post-training, historical context tokens $\mathbf{x}_{t-p:t}$ serve as unnoised conditional anchors $\mathbf{c}_\text{mem}$. The target future tokens $\mathbf{x}_k$, $k \in \{t, \dots, t+q\}$, are used to construct noised latent variables via linear interpolation with Gaussian noise $\boldsymbol{\epsilon} \sim \mathcal{N}(\mathbf{0}, \mathbf{I})$ under noise scale $\lambda \in [0, 1]$:
\begin{equation}
    \mathbf{z}_{\lambda,k} = (1 - \lambda) \mathbf{x}_k + \lambda \boldsymbol{\epsilon}.
\end{equation}
A Video Diffusion Transformer (DiT) $\phi_\theta$ is trained to predict the velocity field that directs the flow from noise back to the data distribution. The model is optimized by minimizing the Mean Squared Error (MSE) of the predicted flow:
\begin{equation}
    \mathcal{L} = \mathbb{E}_{\lambda, \mathbf{x}, \boldsymbol{\epsilon}, \mathbf{c}} \left[ w(\lambda) \| \phi_\theta(\mathbf{z_\lambda}, \lambda, \mathbf{c}) - (\boldsymbol{\epsilon} - \mathbf{x}) \|^2_2 \right],
\end{equation}
where $w(\lambda)$ is a weighting function that assigns adaptive importance to the velocity prediction loss for stable training~\cite{esser2024scalingrectifiedflowtransformers} and $(\boldsymbol{\epsilon} - \mathbf{x})$ denotes the ground-truth target velocity. The conditioning vector $\mathbf{c} = \{ \mathbf{c}_{\text{mem}}, \mathbf{c}_{\text{geo}} \}$ integrates historical context conditions $\mathbf{c}_{\text{mem}}$ and dense geometric action conditions $\mathbf{c}_{\text{geo}}$.





\subsection{World Model for Data Aggregation}
\begin{figure}[ht]   
    \centering
    \includegraphics[width=0.45\textwidth]{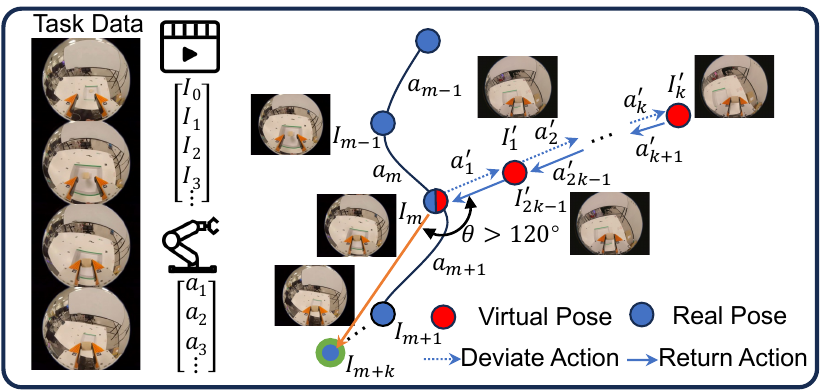}
    \caption{Corrective Action Synthesis Module.}
    \label{fig:virtual_traj} 
    \vspace{-1em}
\end{figure}

With a trained EAC-WM, we now conduct data aggregation for imitation learning. First, we use the Corrective Action Synthesis Module to derive recovery actions, and then use the Consistency-Guided Filtering Module to filter the physically and visually inconsistent generated data.

\subsubsection{Corrective Action Synthesis Module}


This module leverages EAC-WM to synthesize recovery data for OOD states that deviate from the expert demonstrations.
As shown in Figure~\ref{fig:virtual_traj}, for an expert demonstration $\tau = \{\mathbf{a}_i, \mathbf{I}_i\}_{i=1}^n \in \mathcal{D}$, a pivot timestep $m$ is randomly selected as the anchor for synthesis. The deviation horizon is defined as $k$, and a random unit vector $\mathbf{v}_d \in \mathbb{R}^3$ is sampled to represent the direction of a potential OOD state. To avoid misleading supervision, directions are filtered where the angle between $\mathbf{v}_d$ and the subsequent expert action $\mathbf{a}_{m+1}$ is less than $120^\circ$. This constraint ensures that synthesized recovery actions do not oppose the direction of expert trajectory, thereby preventing contradictory training signals that could lead to policy divergence. To maintain the speed of action consistent with the expert demonstration, the deviation displacement of each action in the synthesized recovery data is set to the average action displacement across Task Data $\mathcal{D}$.

The synthesized trajectory $\tau' = \{\mathbf{a}'_j, \hat{\mathbf{I}}_j\}_{j=1}^{2k}$ is constructed in two symmetric phases. First, in the \textit{Deviation Phase} ($\tau'_d$), the robot is steered from the expert pose at timestep $m$ to a perturbed OOD state along $\mathbf{v}_d$. Second, in the \textit{Recovery Phase} ($\tau'_r$), the robot returns from the perturbed state back to the original expert manifold at $m$. Two phases follow the direction of $\mathbf{v}_d$ and $-\mathbf{v}_d$, respectively. 
Given the historical visual states $\mathbf{I}_{m-p:m}$ and the synthesized action sequence $\mathbf{a}'_{1:2k}$ as conditions, the world model predicts the visual states of the synthesized trajectory:
\begin{equation}
    \hat{\mathbf{I}}_{1:2k} = f_\theta \left( \mathbf{I}_{m-p:m}, \mathbf{a}'_{1:2k} \right).
\end{equation}

\begin{figure}[ht]   
    \centering
    \includegraphics[width=0.4\textwidth]{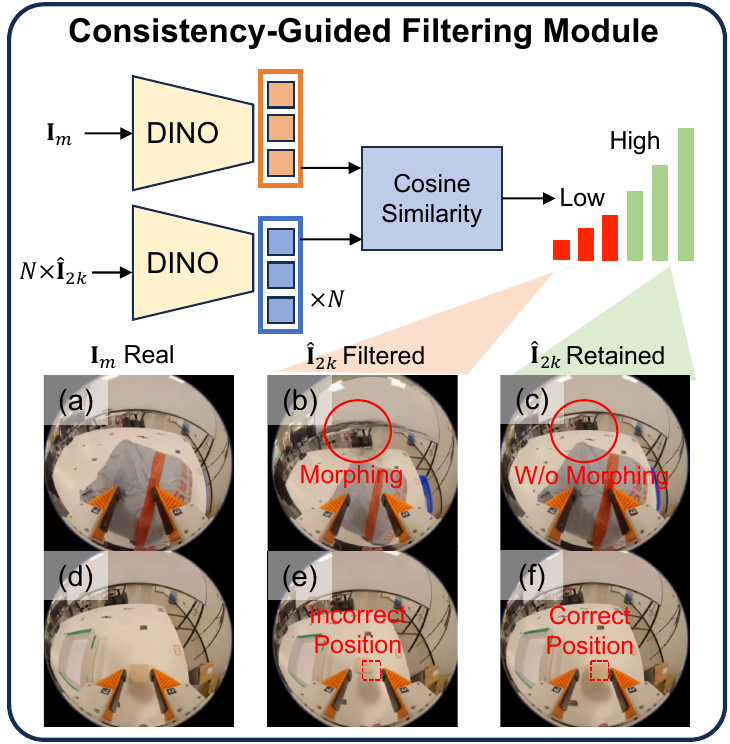}
    \caption{Consistency-Guided Filtering Module and Visualization of Real, Filtered, and Retained Frames. Frames with morphing and incorrect object position are filtered.}
    \label{fig:tsne} 
    \vspace{-1em}
\end{figure}

\begin{figure*}[t]   
    \centering
    \includegraphics[width=1\textwidth]{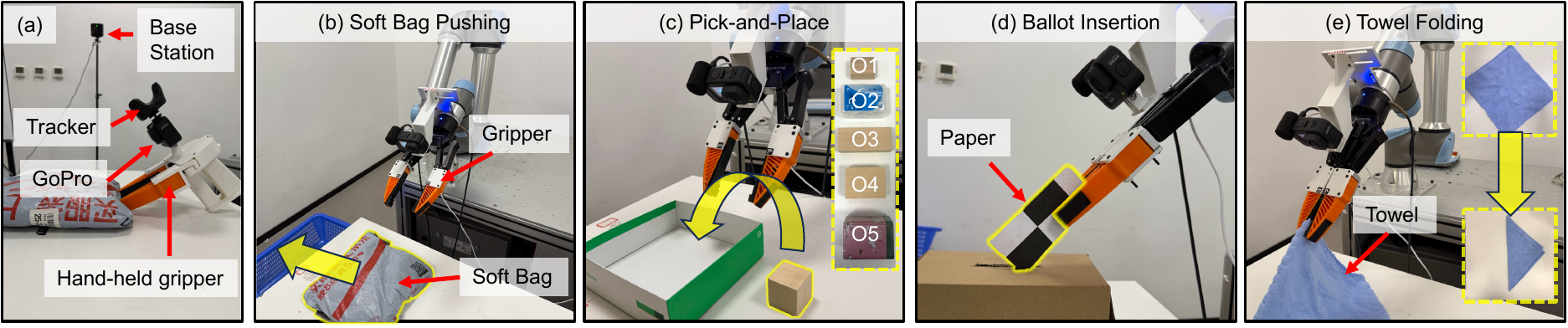}
    \caption{Experimental setup and manipulation tasks. (a) Hardware configuration for data collection, featuring a handheld gripper integrated with a GoPro camera for visual observations and an HTC Vive Tracker for high-precision 6-DoF trajectory tracking. (b--e) The four evaluation tasks: (b) Soft Bag Pushing, (c) Pick-and-Place, (d) Ballot Insertion, and (e) Towel Folding.}
    \label{fig:exp-setup} 
    \vspace{-1em}
\end{figure*}

For a whole synthesized trajectory $\tau'$, we discard the Deviation Phase $\tau'_d$, and only the Recovery Phase $\tau'_r$ is retained for policy training. This ensures that the model specifically learns to recover from OOD states.


\subsubsection{Consistency-Guided Filtering Module}
WMs inevitably produce hallucinations, such as object morphing or physically inconsistent object states. To ensure data fidelity, we introduce a Consistency-Guided Filtering Module to identify and discard these samples.
Our insight is grounded in the temporal error accumulation inherent to WMs. Since the terminal frame $\hat{\mathbf{I}}_{2k}$ is temporally most distant from the conditioning frames, it holds the maximum potential hallucination~\cite{zhao2025ultravico}. Consequently, it can serve as a rigorous proxy for the quality of the entire rollout: if $\hat{\mathbf{I}}_{2k}$ remains consistent with the expert observation $\mathbf{I}_{m}$ which is at the same viewing location, we can reliably infer that the synthesized frames maintain high physical and visual integrity.

To quantify the visual hallucination, we utilize a pre-trained DINOv2~\cite{oquab2024dinov2learningrobustvisual} encoder to extract embedding features of both frames and then calculate their cosine similarity. We employ an adaptive thresholding strategy, filtering out synthesized trajectories with below-average similarity scores to discard hallucinated samples.

Figure~\ref{fig:tsne} illustrates the filtering process. Figures~\ref{fig:tsne}(a) and (d) show the expert demonstration frames for soft bag pushing and pick-and-place, respectively. Figure~\ref{fig:tsne}(b) shows a discarded frame where image morphing occurs, while Figure~\ref{fig:tsne}(e) shows a discarded frame where the block's location deviates from the expert reference. In contrast, Figures~\ref{fig:tsne}(c) and (f) are retained frames, where no significant hallucination is observed.
 


\subsection{Policy Training}
\label{subsec:policy_training}

The training set, $\mathcal{D}_{\text{aug}} = \mathcal{D} \cup \mathcal{D}_{\text{virtual}}$, is constructed by aggregating the expert demonstrations $\mathcal{D}$ with the synthesized trajectories $\mathcal{D}_{\text{virtual}}$. 
To promote temporal consistency of actions, we adopt the action chunking paradigm~\cite{fu2024mobile}. The policy is formulated to predict a sequence of future actions over a horizon $H$ given the observation $\mathbf{I}_t$:
$
\hat{\mathbf{A}}_t = \pi(\mathbf{I}_t) = [\hat{\mathbf{a}}_t, \hat{\mathbf{a}}_{t+1}, \dots, \hat{\mathbf{a}}_{t+H-1}]
$.
The policy is trained to minimize the Mean Squared Error between the predicted chunk and actions in the training set:
\begin{equation}
\mathcal{L}_{\text{policy}} = \mathbb{E}_{(\mathbf{I}_t, \mathbf{A}_t) \sim \mathcal{D}_{\text{aug}}} \left[ \frac{1}{H} \sum_{i=0}^{H-1} \left\| \hat{\mathbf{a}}_{t+i} - \mathbf{a}_{t+i} \right\|_2^2 \right].
\end{equation}
By training on $\mathcal{D}_{\text{aug}}$, the policy internalizes both the expert behaviors and the recovery ability under OOD states.

\begin{figure*}[ht]   
    \centering
    \includegraphics[width=\textwidth]{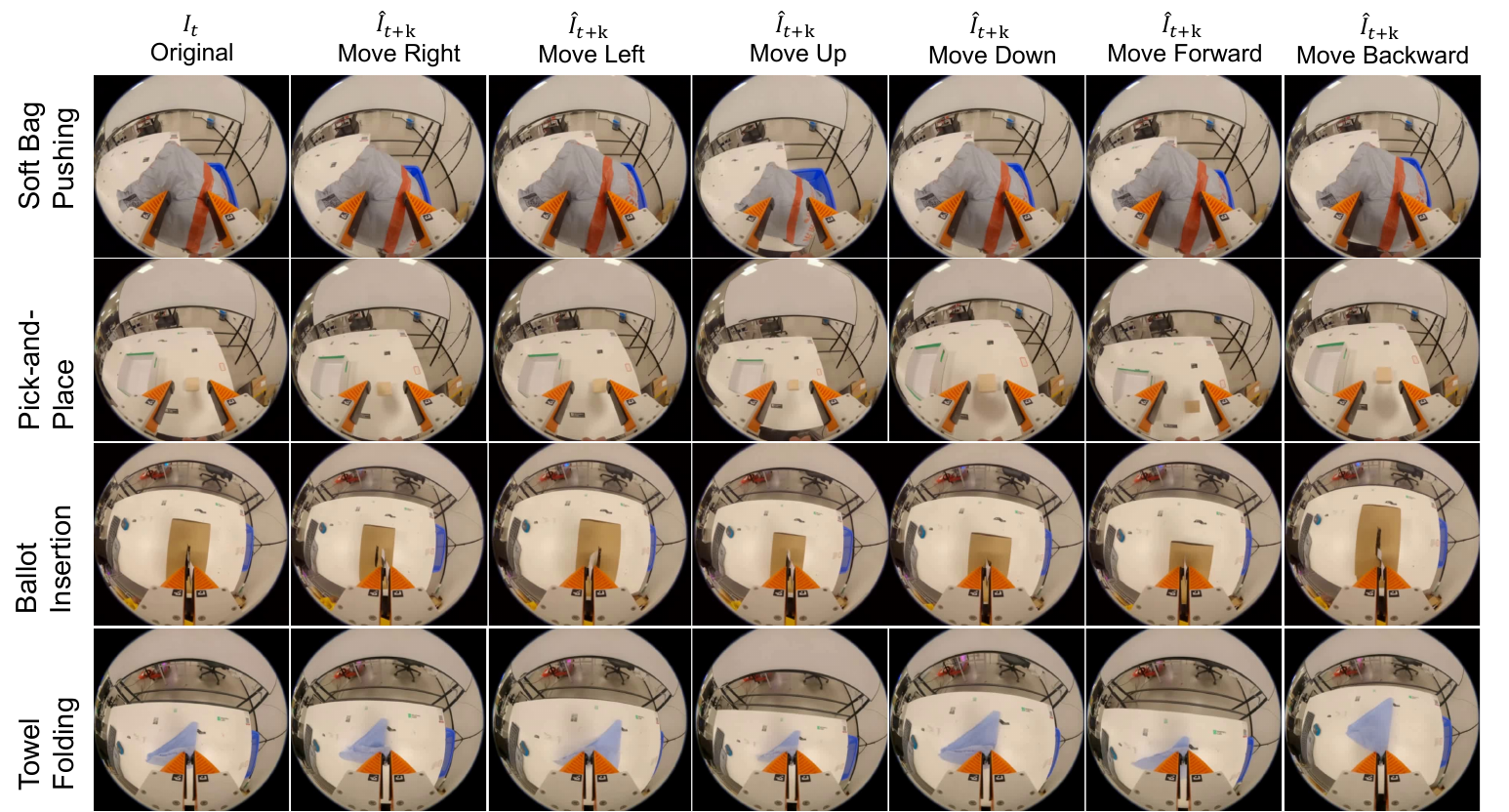}
    \caption{Visualization results of EAC-WM-generated frames. Each row depicts a specific task, starting from the real expert frame $I_t$ (first column). Subsequent columns ($\hat{I}_{t+k}$) show synthesized observations conditioned on action sequences in six cardinal directions (right, left, up, down, forward, and backward).}
    \label{fig:6dir} 
    \vspace{-1em}
\end{figure*}

\section{Evaluation}


We evaluate our framework across four tasks, as shown in Figure~\ref{fig:exp-setup}(b--e): soft bag pushing, pick-and-place, ballot insertion, and towel folding. These tasks test the framework across multiple dimensions, including 
6-DoF 
action spaces, rigid and deformable object manipulation, generalization to novel object instances (pick-and-place, pushing), and high-precision, contact-rich control (insertion). 

On the soft bag pushing, we present visual comparisons of EAC-WM against the baseline generative model. We provide comparisons of success rate with baselines across all tasks, and ablation studies using soft bag pushing to justify our design choices.
The subsequent subsections detail our experiment setups and provide experimental results.

\subsection{Experiment Setups}


\subsubsection{Hardware Platform}

The data collection pipeline largely follows the Universal Manipulation Interface (UMI) framework~\cite{chi2024universal}. As shown in Figure~\ref{fig:exp-setup}(a), we utilize a handheld two-finger gripper equipped with an eye-in-hand GoPro camera featuring a fisheye lens as the visual observation source. 
For demonstration pose collection, we capture the gripper's 6-DoF pose using HTC Vive~\cite{lwowski2020htc}. 
The robotic system includes a Universal Robots UR7e robotic arm and a Robotiq 2F-140 gripper with shark-fin fingertips. Models are trained on 4 and inferred on 1 NVIDIA L20 GPU.

\subsubsection{World Model and Policy Architectures}

We adopt EAC-WM as the generative world model, which is constructed on top of GE-Sim~\cite{liao2025genie} and takes Cosmos-Predict2.5 (2B)~\cite{agarwal2025cosmos} as its foundation model to synthesize high-fidelity, action-conditioned future observations. For the robotic policy, we adopt Gr00t N1.5~\cite{gr00tn1_2025}, a Vision-Language-Action model that serves as the policy model of our framework.

\begin{figure}[ht]   
    \centering
    \includegraphics[width=0.45\textwidth]{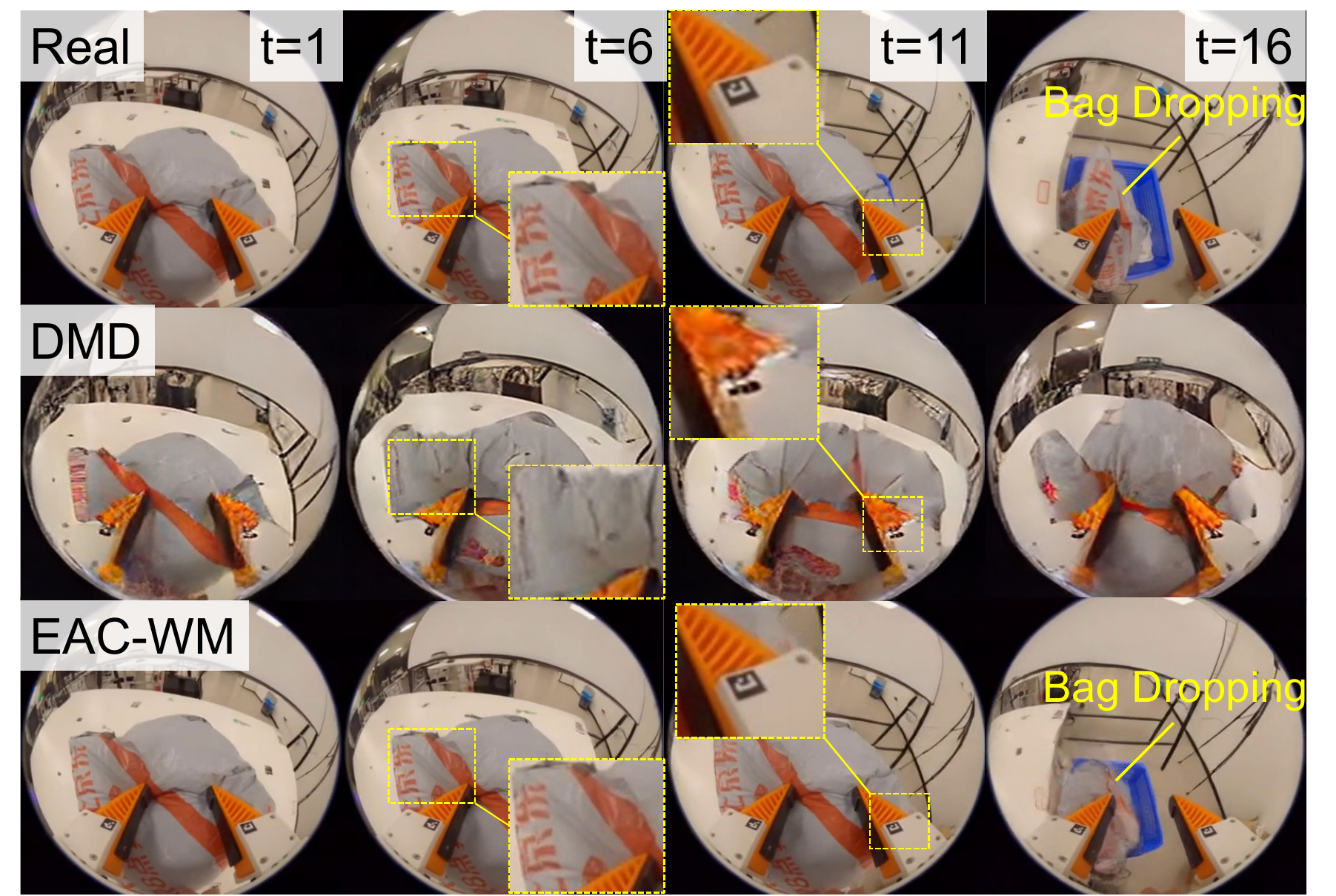}
    \caption{Visual evaluation of EAC-WM versus DMD in generating soft bag pushing data. EAC-WM exhibits superior structural and physical consistency. Notably, at t=16, EAC-WM accurately simulates the dynamics of the bag dropping.}
    \label{fig:wm_dmd} 
    \vspace{-1em}
\end{figure}

\subsubsection{Baselines}
We compare WM-DAgger against two baselines:
\textbf{(1) Standard Behavioral Cloning (BC)}, a standard imitation learning baseline trained without data aggregation. It provides a reference for the performance degradation caused by OOD during deployment.
\textbf{(2) Diffusion Meets DAgger (DMD)}~\cite{zhang2024diffusion}, a SOTA method using diffusion-based synthesis for Data Aggregation. It serves as a baseline to demonstrate that WMs provide superior action-conditional physical consistency compared to standard generative augmentation approaches.

\subsubsection{Data Preparation}
We collect 5 minutes of Play Data for each task. During Play Data collection, the demonstrator holds the gripper, randomly moves around the environment, and interacts with objects. During Task Data collection, we collect 20 human demonstrations for each task. We generate 1500 episodes of recovery actions per task by default.

\subsection{Visualization of Generated Data}

Figure~\ref{fig:6dir} visualizes results synthesized by EAC-WM. Starting from an original real frame $I_t$, EAC-WM generates synthesized observations $\hat{I}_{t+k}$ conditioned on action sequences in six distinct spatial directions. As shown in the Figure, EAC-WM maintains high visual fidelity and physical consistency. It captures complex physical dynamics, including the deformation of the soft bag during pushing and the intricate wrinkling of the towel during folding.

Furthermore, Figure~\ref{fig:wm_dmd} presents a comparison between EAC-WM and DMD. The top row displays the real frames from the expert demonstration, while the second and third rows show the synthesized frames. Both models are conditioned with the same action sequences to predict visual states at timesteps $t=1$, $6$, $11$, and $16$. The results reveal that while DMD suffers from severe visual morphing, EAC-WM consistently preserves both structural and physical integrity of the frames. Notably, EAC-WM realistically simulates the complex physics of the bag dropping at $t=16$.

\subsection{Task 1: Soft Bag Pushing}

\subsubsection{Task Description}

This task requires the robot to push a deformable soft bag across the workspace and deposit it into a target basket. This setup mirrors typical automated sorting scenarios in logistics. During data collection, the initial poses of the bag are randomized. For online validation, we evaluate the policy across ten bag locations, each with three trials, totaling 30 runs.

\begin{table}[t]
\centering
\caption{Success Rates(\%) for Task 1: Soft Bag Pushing across varying numbers of expert demonstrations.}
\label{tab:task1_results}
\begin{tabular}{lccccc}
\toprule
\textbf{Method} & \textbf{1-shot} & \textbf{3-shot} & \textbf{5-shot} & \textbf{10-shot} & \textbf{20-shot} \\
\midrule
Standard BC & 6.7 & 20.0 & 26.7 & 30.0 & 30.0 \\
DMD~\cite{zhang2024diffusion} & 13.3 & 33.3 & 40.0 & 53.3 & 56.7 \\
\textbf{Ours} & \textbf{73.3} & \textbf{86.7} & \textbf{93.3} & \textbf{93.3} & \textbf{96.7} \\
\bottomrule
\end{tabular}
\vspace{-1em}
\end{table}

\subsubsection{Overall Results} 

Table~\ref{tab:task1_results} presents the success rates across varying numbers of expert demonstrations. Standard BC severely struggles with compounding errors, with only 30.0\% success rate even with 20 demonstrations. A typical failure mode involves the end-effector deviating from the correct trajectory and drifting into unrecoverable states, illustrating the classic phenomenon of compounding execution errors. While DMD improves the 20-shot performance to 56.7\%, it remains bottlenecked by a lack of action-conditioned physical consistency in the synthesized frames. Remarkably, WM-DAgger achieves a 73.3\% success rate with only 1-shot demonstration. As demonstrations increase to 20, WM-DAgger reaches 96.7\% success rate. These results validate that synthesizing physically consistent, goal-directed recovery trajectories via WMs effectively teaches the policy to recover from OOD states, thereby acting robustly with minimal human demonstration.

    

\begin{table}[htbp]
\centering
\small
\caption{Policy Performance vs. Data Scaling.}
\label{tab:scaling_horizontal}
\begin{tabular}{lcccc}
\toprule
\textbf{Synthesized Samples} & 300 & 900 & 1500 & 3000 \\
\midrule
\textbf{Success Rate (\%)} & 46.7 & 63.3 & 96.7 & 96.7 \\
\bottomrule
\end{tabular}
\vspace{-1em}
\end{table}

\subsubsection{Impact of Scaling Generated Data}

We then study the impact of scaling WM-DAgger generated data for policy training. As summarized in Table~\ref{tab:scaling_horizontal}, the policy's success rate exhibits a positive correlation with data scaling, as a larger synthetic dataset provides denser coverage of the OOD manifold. 
Notably, while performance improves significantly during the initial scaling phase, the gains begin to marginalize beyond 1500 samples, suggesting that a moderate amount of synthetic data is sufficient for the policy to effectively learn the OOD-recovery behaviors.

\begin{table}[!htbp]
    \centering
    \caption{Ablation Studies for 20-shot Soft Bag Pushing.}
    \label{tab:combined_ablations}
    \begin{tabular}{lc}
        \toprule
        \textbf{Method / Variant} & \textbf{Success Rate (\%)} \\
        \midrule
        \textbf{WM-DAgger (Ours Full)} & \textbf{96.7} \\
        \midrule
        \multicolumn{2}{l}{\textit{Data Ablation}} \\
        \quad w/o Play Data & 83.3 \\
        \midrule
        \multicolumn{2}{l}{\textit{Module Ablation}} \\
        \quad w/o Filter & 66.7 \\
        \quad w/o Dir. & 0.0 \\
        \bottomrule
    \end{tabular}
    \vspace{-1em}
\end{table}

\subsubsection{Impact of Play Data}
Table~\ref{tab:combined_ablations} illustrates the impact of play data in fine-tuning the world model before synthesizing recovery data. When play data is excluded, the policy's success rate drops from 96.7\% to 83.3\%. This decline demonstrates that exposing the world model to diverse environmental interactions is helpful for internalizing physical dynamics for synthesizing higher quality data.

\subsubsection{Module Ablation Studies} 
We conduct ablation studies to evaluate the individual contributions of our designs, as summarized in Table~\ref{tab:combined_ablations}. First, we assess the impact of the Corrective Action Synthesis Module by removing the directional constraint (w/o Dir.). This omission leads to complete task failure (0.0\% success rate). The observed failure mode involves the robot executing erratic movements in arbitrary directions rather than orienting toward the soft bag. This demonstrates that unconstrained recovery actions provide misleading supervision, driving the policy away from the task objective and resulting in collapse of performance.

Second, we evaluate the Consistency-Guided Filtering Module by bypassing the filtering and retaining all generated trajectories (w/o Filter). Without this mechanism to discard visually and physically inconsistent frames, the success rate degrades to 46.7\%. This performance drop confirms that unfiltered hallucinations can introduce biases into the training set, limiting the effectiveness of data aggregation.



\begin{table}[htbp]
\centering
\caption{Success Rates(\%) for Task 2: Pick-and-Place.}
\label{tab:task2_results}
\setlength{\tabcolsep}{3.5pt} 
\begin{tabular}{lccccc}
\toprule
& \multicolumn{3}{c}{\textbf{Seen}} & \multicolumn{2}{c}{\textbf{Unseen}} \\
\cmidrule(lr){2-4} \cmidrule(lr){5-6}
\textbf{Method} & \textbf{O1} & \textbf{O2} & \textbf{O3} & \textbf{O4} & \textbf{O5} \\
\midrule
Standard BC & 13.3 & 13.3 & 6.7 & 0.0 & 10.0 \\
DMD & 33.3 & 36.7 & 26.7 & 6.8 & 16.7 \\
\textbf{Ours} & \textbf{83.3} & \textbf{90.0} & \textbf{80.0} & \textbf{63.3} & \textbf{76.7} \\
\bottomrule
\end{tabular}%
\vspace{-1em}
\end{table}

\subsection{Task 2: Pick-and-Place}

\subsubsection{Task Description}
This task requires reaching, grasping, and depositing a rigid object into a target box. We collect 20 expert demonstrations for each object. During online validation, we conduct 30 trials for each object. The evaluation comprises three \textit{seen} objects (present in the training data, and used for data aggregation) and two \textit{unseen} objects (not used in both training and data aggregation), allowing us to assess the system's visual generalization capabilities.

\subsubsection{Online Validation Results}
Table~\ref{tab:task2_results} presents the task success rates. Even on the seen objects (O1--3), Standard BC frequently fails due to compounding errors during the reaching and grasping phases. A frequent failure mode occurs when the gripper approaches the target but subsequently drifts away and fails to recover. In contrast, WM-DAgger achieves much higher success rates. By synthesizing recovery trajectories in OOD states, our framework equips the policy with robust recovery abilities.

\subsubsection{Generalizability across Unseen Objects}
When evaluating on unseen objects (O4--5), standard BC suffers from low success rates (0.0\% and 10.0\%). Conversely, WM-DAgger maintains moderate success rates (63.3\% and 76.7\%). By synthesizing diverse visual and physical states during the data aggregation phase, WM-DAgger prevents the policy from memorizing specific visual textures, enhancing generalizability across unseen objects.

\begin{table}[htbp]
    \small 
    \centering

    \begin{minipage}[t]{0.48\columnwidth}
        \centering
        \caption{Success Rates for Ballot Insertion.}
        \label{tab:task3_results}
        \begin{tabular}{lc}
            \toprule
            \textbf{Method} & \textbf{Succ. (\%)} \\
            \midrule
            Standard BC & 13.3 \\
            DMD & 26.7 \\
            \textbf{Ours} & \textbf{73.3} \\
            \bottomrule
        \end{tabular}
    \end{minipage}
    \hfill 
    \begin{minipage}[t]{0.48\columnwidth}
        \centering
        \caption{Success Rates for Towel Folding.}
        \label{tab:task4_results}
        \begin{tabular}{lc}
            \toprule
            \textbf{Method} & \textbf{Succ. (\%)} \\
            \midrule
            Standard BC & 0.0 \\
            DMD & 10.0 \\
            \textbf{Ours} & \textbf{46.7} \\
            \bottomrule
        \end{tabular}
    \end{minipage}
    \vspace{-1em}
\end{table}

\subsection{Task 3: Ballot Insertion}

\subsubsection{Task Description}
The robot is required to grasp a deformable ballot paper and insert it into a narrow slot of a ballot box. It is highly challenging as it couples the manipulation of a deformable planar object with a high-precision, contact-rich insertion phase. We utilize this task to validate the capability of our framework to support tasks requiring high precision and complex physical interactions.

\subsubsection{Online Validation Results}
Table~\ref{tab:task3_results} presents the results. Standard BC struggles significantly, achieving only 13.3\% success rate. As the insertion slot is narrow, minor execution drift could cause the ballot to miss the target or crumple against the box. In contrast, WM-DAgger achieves a much higher success rate of 73.3\%. By synthesizing massive recovery actions, our method equips the policy with the closed-loop adjustment capabilities to re-align the ballot and successfully complete the high-precision insertion.

\subsection{Task 4: Towel Folding}

\subsubsection{Task Description}
This task requires the robot to grasp and fold a deformable towel. Towel folding demands complex 6-DoF spatial trajectories. This experiment evaluates the framework's scalability to more complex action spaces and severe morphological variations of the object. 

\subsubsection{Online Validation Results}
The results are presented in Table~\ref{tab:task4_results}. Operating in a 6-DoF action space exacerbates the compounding error problem for Standard BC, as errors in both position and orientation could accumulate, dropping success rate to 0.0\%. For DMD, as generating physically consistent towel dynamics is difficult, the success rate is still low (10\%). However, WM-DAgger achieves a success rate of 46.7\%. The EAC-WM effectively simulates the deformable towel under 6-DoF manipulations, generating massive physically grounded recovery data for policy training. This allows the policy to complete the task more robustly.
\section{Conclusion and Future Work}

This paper presents WM-DAgger, a novel framework that leverages World Models to synthesize physically consistent recovery data for imitation learning. By integrating the Corrective Action Synthesis Module and the Consistency-Guided Filtering Module, WM-DAgger enables policies to learn robust error-recovery behaviors, bypassing the need for expensive extra human demonstrations. Extensive real-world validations confirm significant improvements in success rates of four diverse tasks. Our results demonstrate that World Models can serve as scalable, high-fidelity supervisors for developing reliable embodied intelligence.

While WM-DAgger demonstrates strong performance with two-finger grippers, scaling to dexterous multi-finger hands remains a challenge. The high DoF inherent in dexterous manipulation complicate the synthesis of visually and physically consistent articulated frames. In future work, we plan to address this by integrating morphological priors and kinematic topologies into the world model conditioning to support high-DoF dexterous manipulation.

\addtolength{\textheight}{-12cm}   


\bibliographystyle{IEEEtran}
\bibliography{References}

@misc{esser2024scalingrectifiedflowtransformers,
      title={Scaling Rectified Flow Transformers for High-Resolution Image Synthesis}, 
      author={Patrick Esser and Sumith Kulal and Andreas Blattmann and Rahim Entezari and Jonas Müller and Harry Saini and others},
      year={2024},
      eprint={2403.03206},
      archivePrefix={arXiv},
      primaryClass={cs.CV},
      url={https://arxiv.org/abs/2403.03206}, 
}

@inproceedings{kelly2019hg,
  title={Hg-dagger: Interactive imitation learning with human experts},
  author={Kelly, Michael and Sidrane, Chelsea and Driggs-Campbell, Katherine and Kochenderfer, Mykel J},
  booktitle={2019 International Conference on Robotics and Automation (ICRA)},
  pages={8077--8083},
  year={2019},
  organization={IEEE}
}

@inproceedings{ross2011reduction,
  title={A reduction of imitation learning and structured prediction to no-regret online learning},
  author={Ross, St{\'e}phane and Gordon, Geoffrey and Bagnell, Drew},
  booktitle={Proceedings of the fourteenth international conference on artificial intelligence and statistics},
  pages={627--635},
  year={2011},
  organization={JMLR Workshop and Conference Proceedings}
}

@article{schaal1999imitation,
  title={Is imitation learning the route to humanoid robots?},
  author={Schaal, Stefan},
  journal={Trends in cognitive sciences},
  volume={3},
  number={6},
  pages={233--242},
  year={1999},
  publisher={Elsevier}
}

@article{bai2025towards,
  title={Towards a unified understanding of robot manipulation: A comprehensive survey},
  author={Bai, Shuanghao and Song, Wenxuan and Chen, Jiayi and Ji, Yuheng and Zhong, Zhide and Yang, Jin and Zhao, Han and Zhou, Wanqi and Zhao, Wei and Li, Zhe and others},
  journal={arXiv preprint arXiv:2510.10903},
  year={2025}
}

@article{zhang2024diffusion,
  title={Diffusion meets dagger: Supercharging eye-in-hand imitation learning},
  author={Zhang, Xiaoyu and Chang, Matthew and Kumar, Pranav and Gupta, Saurabh},
  journal={arXiv preprint arXiv:2402.17768},
  year={2024}
}

@article{ding2025understanding,
  title={Understanding world or predicting future? a comprehensive survey of world models},
  author={Ding, Jingtao and Zhang, Yunke and Shang, Yu and Zhang, Yuheng and Zong, Zefang and Feng, Jie and Yuan, Yuan and Su, Hongyuan and Li, Nian and Sukiennik, Nicholas and others},
  journal={ACM Computing Surveys},
  volume={58},
  number={3},
  pages={1--38},
  year={2025},
  publisher={ACM New York, NY}
}

@article{hafner2023mastering,
  title={Mastering diverse domains through world models},
  author={Hafner, Danijar and Pasukonis, Jurgis and Ba, Jimmy and Lillicrap, Timothy},
  journal={arXiv preprint arXiv:2301.04104},
  year={2023}
}

@article{zare2024survey,
  title={A survey of imitation learning: Algorithms, recent developments, and challenges},
  author={Zare, Maryam and Kebria, Parham M and Khosravi, Abbas and Nahavandi, Saeid},
  journal={IEEE Transactions on Cybernetics},
  volume={54},
  number={12},
  pages={7173--7186},
  year={2024},
  publisher={IEEE}
}

@article{xu2025compliant,
  title={Compliant residual dagger: Improving real-world contact-rich manipulation with human corrections},
  author={Xu, Xiaomeng and Hou, Yifan and Xin, Chendong and Liu, Zeyi and Song, Shuran},
  journal={arXiv preprint arXiv:2506.16685},
  year={2025}
}

@inproceedings{yu2025manigaussian++,
  title={ManiGaussian++: General robotic bimanual manipulation with hierarchical Gaussian world model},
  author={Yu, Tengbo and Lu, Guanxing and Yang, Zaijia and Deng, Haoyuan and Chen, Season Si and Lu, Jiwen and Ding, Wenbo and Hu, Guoqiang and Tang, Yansong and Wang, Ziwei},
  booktitle={2025 IEEE/RSJ International Conference on Intelligent Robots and Systems (IROS)},
  pages={12232--12239},
  year={2025},
  organization={IEEE}
}

@inproceedings{li2025simworld,
  title={Simworld: A unified benchmark for simulator-conditioned scene generation via world model},
  author={Li, Xinqing and Song, Ruiqi and Xie, Qingyu and Wu, Ye and Zeng, Nanxin and Ai, Yunfeng},
  booktitle={2025 IEEE/RSJ International Conference on Intelligent Robots and Systems (IROS)},
  pages={927--934},
  year={2025},
  organization={IEEE}
}

@article{agarwal2025cosmos,
  title={Cosmos world foundation model platform for physical ai},
  author={Agarwal, Niket and Ali, Arslan and Bala, Maciej and Balaji, Yogesh and Barker, Erik and Cai, Tiffany and Chattopadhyay, Prithvijit and Chen, Yongxin and Cui, Yin and Ding, Yifan and others},
  journal={arXiv preprint arXiv:2501.03575},
  year={2025}
}

@inproceedings{wu2023daydreamer,
  title={Daydreamer: World models for physical robot learning},
  author={Wu, Philipp and Escontrela, Alejandro and Hafner, Danijar and Abbeel, Pieter and Goldberg, Ken},
  booktitle={Conference on robot learning},
  pages={2226--2240},
  year={2023},
  organization={PMLR}
}

@article{jiang2025world4rl,
  title={World4rl: Diffusion world models for policy refinement with reinforcement learning for robotic manipulation},
  author={Jiang, Zhennan and Liu, Kai and Qin, Yuxin and Tian, Shuai and Zheng, Yupeng and Zhou, Mingcai and Yu, Chao and Li, Haoran and Zhao, Dongbin},
  journal={arXiv preprint arXiv:2509.19080},
  year={2025}
}

@article{chi2024universal,
  title={Universal manipulation interface: In-the-wild robot teaching without in-the-wild robots},
  author={Chi, Cheng and Xu, Zhenjia and Pan, Chuer and Cousineau, Eric and Burchfiel, Benjamin and Feng, Siyuan and Tedrake, Russ and Song, Shuran},
  journal={arXiv preprint arXiv:2402.10329},
  year={2024}
}

@article{lwowski2020htc,
  title={HTC Vive tracker: accuracy for indoor localization},
  author={Lwowski, Jonathan and Majumdat, Abhijit and Benavidez, Patrick and Prevost, John J and Jamshidi, Mo},
  journal={IEEE Systems, Man, and Cybernetics Magazine},
  volume={6},
  number={4},
  pages={15--22},
  year={2020},
  publisher={IEEE}
}

@article{fu2024mobile,
  title={Mobile aloha: Learning bimanual mobile manipulation with low-cost whole-body teleoperation},
  author={Fu, Zipeng and Zhao, Tony Z and Finn, Chelsea},
  journal={arXiv preprint arXiv:2401.02117},
  year={2024}
}

@inproceedings{gr00tn1_2025,
  archivePrefix = {arxiv},
  eprint     = {2503.14734},
  title      = {{GR00T} {N1}: An Open Foundation Model for Generalist Humanoid Robots},
  author     = {NVIDIA and Johan Bjorck and Fernando Castañeda, Nikita Cherniadev and Xingye Da and others},
  month      = {March},
  year       = {2025},
  booktitle  = {ArXiv Preprint},
}

@article{liao2025genie,
  title={Genie envisioner: A unified world foundation platform for robotic manipulation},
  author={Liao, Yue and Zhou, Pengfei and Huang, Siyuan and Yang, Donglin and Chen, Shengcong and Jiang, Yuxin and others},
  journal={arXiv preprint arXiv:2508.05635},
  year={2025}
}

@inproceedings{brown2019extrapolating,
  title={Extrapolating beyond suboptimal demonstrations via inverse reinforcement learning from observations},
  author={Brown, Daniel and Goo, Wonjoon and Nagarajan, Prabhat and Niekum, Scott},
  booktitle={International conference on machine learning},
  pages={783--792},
  year={2019},
  organization={PMLR}
}

@article{li2025comprehensive,
  title={A comprehensive survey on world models for embodied ai},
  author={Li, Xinqing and He, Xin and Zhang, Le and Wu, Min and Li, Xiaoli and Liu, Yun},
  journal={arXiv preprint arXiv:2510.16732},
  year={2025}
}

@article{liu2022flow,
  title={Flow straight and fast: Learning to generate and transfer data with rectified flow},
  author={Liu, Xingchao and Gong, Chengyue and Liu, Qiang},
  journal={arXiv preprint arXiv:2209.03003},
  year={2022}
}

@misc{oquab2024dinov2learningrobustvisual,
      title={DINOv2: Learning Robust Visual Features without Supervision}, 
      author={Maxime Oquab and Timothée Darcet and Théo Moutakanni and Huy Vo and Marc Szafraniec and Vasil Khalidov and others},
      year={2024},
      eprint={2304.07193},
      archivePrefix={arXiv},
      primaryClass={cs.CV},
      url={https://arxiv.org/abs/2304.07193}, 
}

@article{zhao2025ultravico,
  title={UltraViCo: Breaking Extrapolation Limits in Video Diffusion Transformers},
  author={Zhao, Min and Zhu, Hongzhou and Wang, Yingze and Yan, Bokai and Zhang, Jintao and He, Guande and Yang, Ling and Li, Chongxuan and Zhu, Jun},
  journal={arXiv preprint arXiv:2511.20123},
  year={2025}
}

@inproceedings{yifan2022input,
  title={Input-level inductive biases for 3D reconstruction},
  author={Yifan, Wang and Doersch, Carl and Arandjelovi{\'c}, Relja and Carreira, Joao and Zisserman, Andrew},
  booktitle={Proceedings of the IEEE/CVF Conference on Computer Vision and Pattern Recognition},
  pages={6176--6186},
  year={2022}
}

\end{document}